\documentclass[10pt,twocolumn,letterpaper]{article}

\usepackage[algorithms]{wacv}
\usepackage{times}
\usepackage{epsfig}
\usepackage{graphicx}
\usepackage{amsmath}
\usepackage{amssymb}
\usepackage{adjustbox}
\usepackage{subcaption}
\usepackage{multirow}
\usepackage{multicol}
\usepackage{booktabs}
\usepackage{stmaryrd}   
\usepackage{url}
\usepackage{array}
\usepackage[dvipsnames,table,xcdraw]{xcolor}
\usepackage{tikz}
\usetikzlibrary{fit}
\usepackage{pifont}

\definecolor{myorange}{rgb}{1, 0.647, 0}
\definecolor{myblue}{rgb}{.118, 0.565, 1}

\def\wacvPaperID{2182} 
\def\confName{WACV}
\def\confYear{2024}

\usepackage[pagebackref,breaklinks,colorlinks]{hyperref}
\usepackage[capitalize]{cleveref}

\begin{document}

\title{INCODE: Implicit Neural Conditioning with Prior Knowledge Embeddings}

\author{
\quad Amirhossein Kazerouni$^{\;1}$ 
\quad Reza Azad$^{\;2}$
\quad Alireza Hosseini$^{\;3}$ 
\quad Dorit Merhof$^{\;4, 5}$ 
\quad Ulas Bagci$^{\;6}$ 
\\
${^1}$ School of Electrical Engineering, Iran University of Science and Technology, Iran\\
${^2}$ Institute of Imaging and Computer Vision, RWTH Aachen University, Germany \\
${^3}$ School of Electrical and Computer Engineering, College of Engineering, University of Tehran, Iran \\
${^4}$ Faculty of Informatics and Data Science, University of Regensburg, Germany \\
${^5}$ Fraunhofer Institute for Digital Medicine MEVIS, Bremen, Germany \\
${^6}$ Department of Radiology, Northwestern University, Chicago, USA \\
{\tt\small \{amirhossein477, rezazad68\}@gmail.com, \{arhosseini77\}@ut.ac.ir} \\
{\tt\small \{dorit.merhof\}@ur.de, \{ulas.bagci\}@northwestern.edu}  \\
\normalsize \url{https://xmindflow.github.io/incode}
}
\maketitle
\thispagestyle{empty}

\begin{abstract}
Implicit Neural Representations (INRs) have revolutionized signal representation by leveraging neural networks to provide continuous and smooth representations of complex data. However, existing INRs face limitations in capturing fine-grained details, handling noise, and adapting to diverse signal types. To address these challenges, we introduce INCODE, a novel approach that enhances the control of the sinusoidal-based activation function in INRs using deep prior knowledge. INCODE comprises a harmonizer network and a composer network, where the harmonizer network dynamically adjusts key parameters of the activation function. Through a task-specific pre-trained model, INCODE adapts the task-specific parameters to optimize the representation process. Our approach not only excels in representation, but also extends its prowess to tackle complex tasks such as audio, image, and 3D shape reconstructions, as well as intricate challenges such as neural radiance fields (NeRFs), and inverse problems, including denoising, super-resolution, inpainting, and CT reconstruction. Through comprehensive experiments, INCODE demonstrates its superiority in terms of robustness, accuracy, quality, and convergence rate, broadening the scope of signal representation. Please visit the project's website for details on the proposed method and access to the code.

\end{abstract}

\vspace{-0.75em}
\section{Introduction}
The realm of signal representation has undergone a significant transformation with the emergence of Implicit Neural Representations (INRs), also known as coordinate-based neural representations. Unlike traditional methods where signal values are discretely stored on coordinate grids, this new approach revolves around training neural networks, specifically Multilayer Perceptrons (MLPs), equipped with continuous nonlinear activation functions. The goal is to approximate the complex relationship between coordinates and their corresponding signal values, ultimately providing a continuous signal representation \cite{sitzmann2020implicit}.

INRs have received considerable attention for their ability to learn tasks involving complex and high-dimensional data more compactly and flexibly. They have shown promise in applications spanning computer graphics \cite{gao2022nerf,muller2022instant,mildenhall2021nerf}, computer vision \cite{maiya2023nirvana,lu2023learning,xu2022signal,molaei2023implicit}, virtual reality \cite{deng2022fov,li2023bringing}, and so on. The inherent attributes of seamlessness and continuity within INRs offer a wide range of advantages, most notably in applications involving super-resolution and inpainting tasks. Unlike Convolutional Neural Networks (CNNs), INRs bypass the limitations attributed to locality biases and leverage the power of neural networks to directly learn the relationship between inputs and desired outputs, thereby enhancing their effectiveness in modeling complex tasks. However, their potential is hampered by limitations. Previous approaches have not fully exploited the high representation capacity of INRs, failing to extract fine-grained details. Additionally, these methods often disregard data noise, rendering them ineffective for tasks such as super-resolution, denoising, and inpainting. Their applicability across signal types is limited, and scalability to handle large signal sets poses difficulties. Overcoming these challenges is crucial for unlocking INRs' efficacy in diverse signal-processing contexts.

Conditional neural networks constitute a significant advancement in deep learning, endowing networks with adaptability based on auxiliary information, a departure from conventional context-agnostic counterparts. This adaptability introduces context awareness and targeted responsiveness. The incorporation of supplementary conditions enables the accommodation of data distribution variations. In the domain of INRs, latent code concatenation with MLP spatial coordinates is prevalent \cite{rebain2022attention,mehta2021modulated,chen2019learning}. An alternative, the dual-MLP approach by Mehta et al. \cite{mehta2021modulated}, deploys a ReLU-based modulator network for amplitude modulation of sinusoidal activations across the hidden layers of the synthesis network. This modulation involves element-wise multiplication of modulator and synthesis activations. Shen et al. \cite{shen2022nerp} augment CT and MRI reconstruction by embedding prior image data into MLP weights, initializing a reconstruction network, and facilitating its training. However, the concatenation strategy imposes limitations on reconstruction quality, the modulated synthesizer approach fails to fully exploit the potential of sinusoidal activation, the utilization of initialization techniques necessitates a two-step process, and using hyper-networks \cite{wu2023hyperinr,klocek2019hypernetwork} is computationally expensive and requires significant memory costs.

To mitigate these problems, we present a novel INR method to enhance the hierarchical representation capabilities of the INRS. The proposed method excels in achieving high-quality reconstructions across various tasks, encompassing audio, image, and 3D shape, as well as intricate challenges such as NeRF and inverse problems including denoising, super-resolution, inpainting, and CT (computed tomography) reconstruction. The architectural foundation of our proposed model is characterized by a dual-component MLP structure, comprising a \textit{harmonizer network} and a \textit{composer network}. The composer network is distinguished by a general form of sinusoidal activation function $(\underline{\textbf{a}}sin(\underline{\textbf{b}} \omega_0 x + \underline{\textbf{c}}) + \underline{\textbf{d}})$, which effectively establishes a mapping between spatial coordinates and their respective values. Concurrently, the harmonizer network conditions the composer with a deep prior knowledge by dynamically adjusting the parameters $\underline{\textbf{a}}$, $\underline{\textbf{b}}$, $\underline{\textbf{c}}$, and $\underline{\textbf{d}}$ during the learning process. To this end, task-specific pre-trained models are used to generate object embeddings. At each learning step, the obtained embedding is fed into the harmonizer network, yielding the extraction of sinusoidal parameters. This symbiotic arrangement empowers the composer network to adeptly capture detailed information and refine intricacies crucial for accurate and comprehensive representation. Furthermore, we employ a regularization technique for the estimated parameters to expedite the convergence of the model. Our extensive experiments on various applications clearly demonstrate the superiority of our approach in terms of robustness, accuracy, quality, and convergence rate.
\begin{figure*}[!t]
    \centering
    \includegraphics[width=0.9\textwidth]{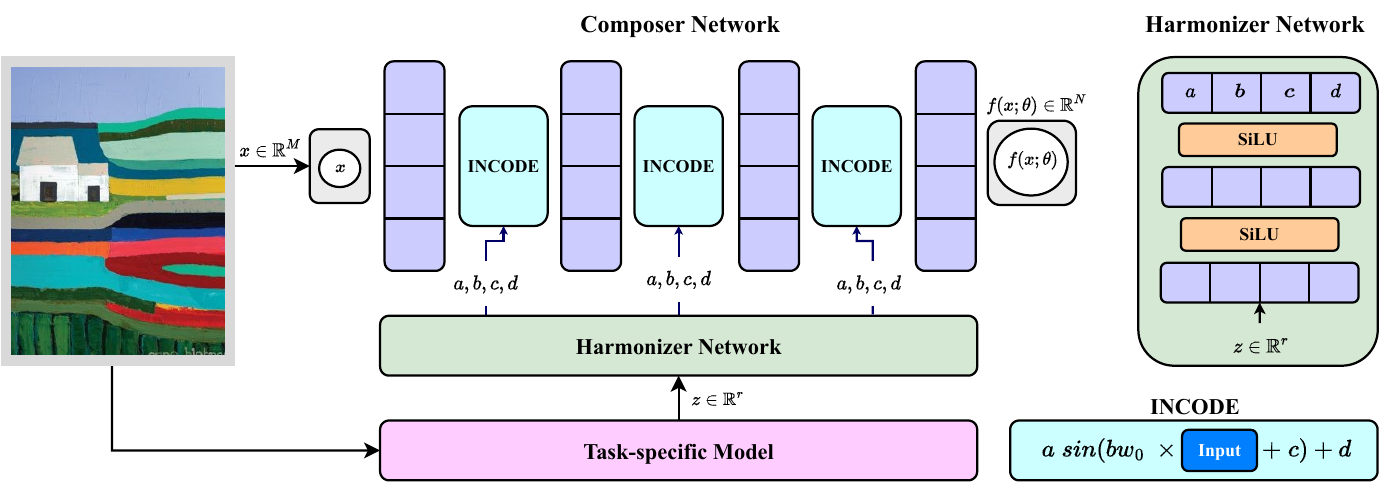}
    \vspace{-0.5em}
    \caption{Illustration of the \textbf{INCODE} pipeline: adaptive implicit neural representation with prior knowledge embedding.}
    \label{fig:Incode}
    \vspace{-0.5em}
\end{figure*}

\section{Related Works}
\noindent\textbf{Implicit Neural Representation.}
Recent works have shown remarkable success in representing various signals using neural networks. INR applications span across several domains: 3D shape, image, video, and audio signals \cite{sitzmann2020implicit,saragadam2023wire,ortiz2022isdf,duggal2022mending, li2022learning,kim2022learning,rho2022neural,strumpler2022implicit}. Most INR works have been pursued to address the challenge of spectral bias encountered in ReLU-based MLPs, which inherently inclines towards learning low-frequency components \cite{rahaman2019spectral}. Sitzmann et al. in \cite{sitzmann2020implicit} utilize sinusoidal-based activation functions for INR. Tanckik et al. \cite{tancik2020fourier} introduce an FFN that applies a Fourier feature mapping before the actual network to promote the learning of high-frequency data. Fathony et al. \cite{fathony2021multiplicative} introduce two variations of the Multiplicative Filter Networks (MFN): one employing sinusoids and another one utilizing a Gabor wavelet as the filter applied after each layer. 
Some works take advantage of using an aggregate of smaller networks to represent the signal rather than using one large MLP. In \cite{kadarvish2021ensemble}, the input signal is broken into regular grids of smaller sizes, and a separate network is responsible for representing each cell inside the grid. \cite{martel2021acorn,saragadam2022miner} introduce an adaptive method for resource allocation based on the local complexity of the signal, enabling INR to work on larger signals, e.g., gigapixel images. Moreover, Mildenhall et al. \cite{mildenhall2020nerf} employ volume rendering to represent 3D scenes that take advantage of coordinate-based neural networks. Since vanilla NeRF is difficult to train and entails lengthy training processes, other methods \cite{chen2023mobilenerf,gao2022nerf,barron2022mip,wang2023f2,li2023steernerf,pumarola2021d} utilized similar approaches to improve the fidelity and efficiency of NeRFs. KiloNeRF \cite{reiser2021kilonerf} shortens the rendering process by three orders of magnitude, where they utilize thousands of tiny MLPs to represent different segments of a scene and merge the outputs to obtain the entire scene. Müller et al. \cite{muller2022instant} have utilized hash encoding to expedite the training and inference process in NeRFs.

\noindent\textbf{Periodic Activation Functions.} 
In recent studies, periodic activation functions have exhibited favorable results in INR tasks by instructing the network to learn high-frequency details. Such activation functions have been widely investigated since 1987, when Lapedes and Farber \cite{lapedes1987nonlinear} showed that networks with such activations are generally difficult to train. Further, Parascandolo et al. \cite{parascandolo2016taming} shed light on why training networks with periodic activation functions is challenging. They show that training is only successful if the networks do not rely on the periodicity of the given functions and propose using a truncated sinusoidal function. Klocek et al. \cite{klocek2019hypernetwork}, motivated by discrete cosine transform, propose to exploit cosine activation functions for a target network whose weights are determined by a hyper-network. Recently, Sitzmann et al. \cite{sitzmann2020implicit} leverage sinusoidal activation functions initialized carefully to represent complex unstructured data. Motivated by this, we propose INCODE, a general form of sinusoidal activation function, aiming to improve the representation accuracy and robustness of SIREN.

\noindent\textbf{Conditional Neural Network.}
In this domain, the focus on improving model adaptability through contextual information integration reflects a broader trend toward harnessing auxiliary data for enhanced model performance. In INRs, a common strategy involves the concatenation of latent codes obtained from an encoder with input coordinates \cite{park2019deepsdf,chen2019learning,rebain2022attention}. Diverse approaches have emerged for contextual integration: Klocek et al. \cite{klocek2019hypernetwork} leverage hyper-networks to compute weights for the primary network operating on coordinates, while Rebain et al. \cite{rebain2022attention} propose an attention MLP conditioning mechanism using the latent code as keys and values and the coordinate as queries. Mehta et al. \cite{mehta2021modulated} modulate the implicit function through a modulator MLP. Consequently, we present a novel conditioning process, wherein we estimate the parameters of the proposed activation function using deep prior information and an auxiliary MLP network, thereby contributing to the growing landscape of adaptive conditional neural networks.

\section{Method}
\vspace{-0.5em}
The INR function operates by encoding a continuous target signal $S(\text{x}): \mathbb{R}^M\mathbb \rightarrow \mathbb{R}^N$ through a neural network $f(\text{x}; \theta): \mathbb{R}^M\mathbb \rightarrow \mathbb{R}^N$, i.e., an MLP, where the network's architecture is parameterized by a set of weights $\theta$. This network establishes a functional mapping between input coordinates $\text{x} \in \mathbb{R}^M$ and signal values $S(\text{x}) \in \mathbb{R}^N$ (e.g., occupancy, color, etc.). This is achieved by minimizing a loss function as:
\begin{equation}
    \underset{\theta}{\text{arg min}}\;\underset{x \in X}{\mathbb{E}} \left [||f(\text{x}; \theta) - S(\text{x})||_{2}^{2}\right].
\end{equation}

By implementing $f(\text{x}; \theta)$ with ReLU-based MLP architectures, a notable trend emerges: the network displays a bias for capturing low-frequency signals. This trait, as shown by Rahaman et al. \cite{rahaman2019spectral}, frequently results in inferior-quality signal reconstructions. Sitzmann et al. \cite{sitzmann2020implicit} propose to use MLP with a sinusoidal activation function (SIREN method), where the post-activation layer is recursively defined as follows:
\begin{equation}
    \text{y}_l = \sin (w_o\left(\text{W}_l \text{y}_{l-1} + \text{b}_{l} \right)), \;\; l=1, 2,...,L-1,
\end{equation}
where $\text{W}_l \in \mathbb{R}^{P_{l-1} \times P_l}$ denotes the weights and $\text{b}_l \in \mathbb{R}^{P_l}$ indicates the bias at the $l_{th}$ layer of the network. While SIREN offers superior representation capacity compared to ReLU, there is still an opportunity to enhance control over the $sine$ activation function. This control could adaptively amplify representation capacity and counter noise's impact, diverging from the original SIREN's noise-equivalent treatment in image representation. Therefore, our focus in \textbf{INCODE} is on introducing a $sine$-based activation function that provides enhanced control throughout the learning process by using deep prior knowledge. 
\subsection{INCODE}
\vspace{-0.5em}
We now present \textbf{INCODE}: a conditional INR model with prior knowledge embeddings, illustrated in \autoref{fig:Incode}.
INCODE is composed of two fundamental components: a \textit{harmonizer} network and a \textit{synthesizer} network. The harmonizer network endeavors to adjust the activation function of the composer network, while the composer network's duty is to craft a final piece. To initiate the process, we obtain a latent code $z \in \mathbb{R}^r$ from a pre-trained model tailored to the task. This latent code then serves as input for the harmonizer network, which conditions the composer network's mapping of spatial coordinates to signal values.
\begin{figure*}[!t]
    \centering
    \includegraphics[width=0.8\textwidth]{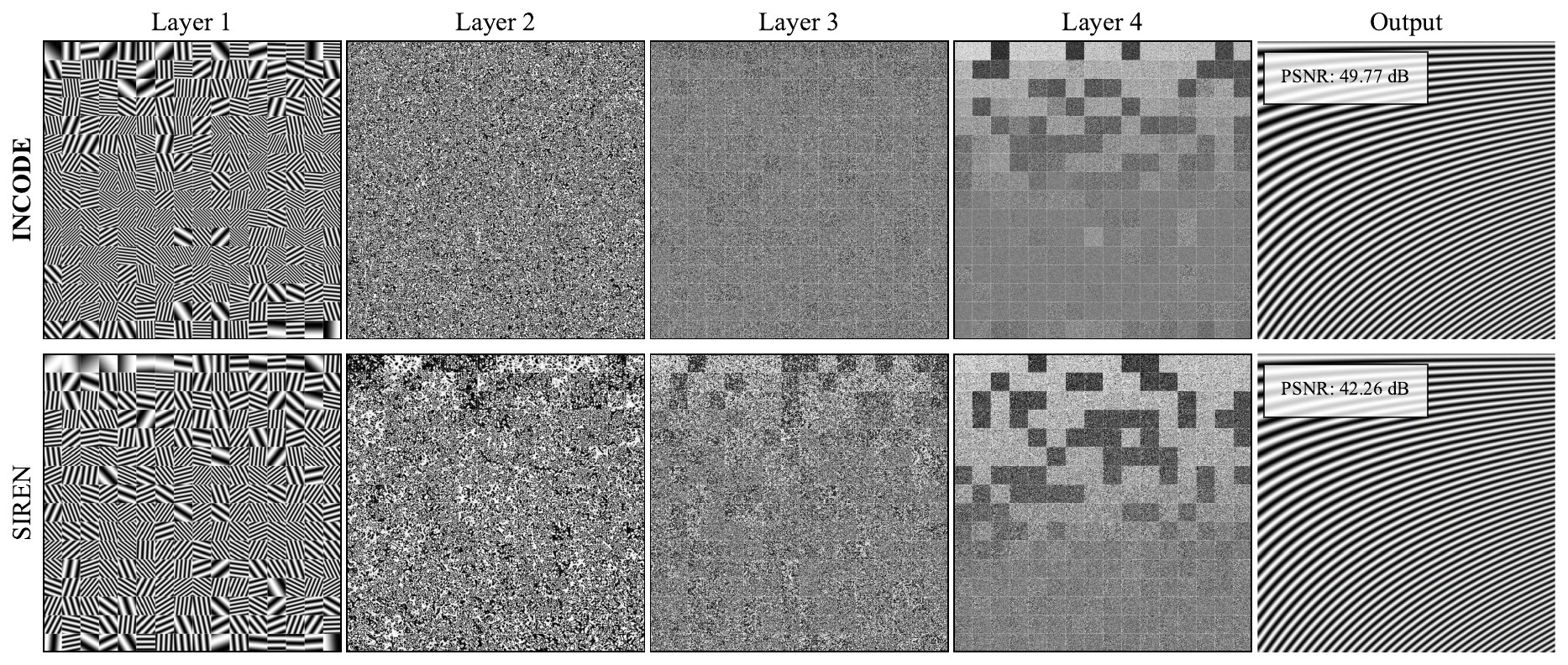}
    \vspace{-1.5em}
\end{figure*}
\begin{figure*}[!t]
    \centering
    \includegraphics[width=0.9\textwidth]{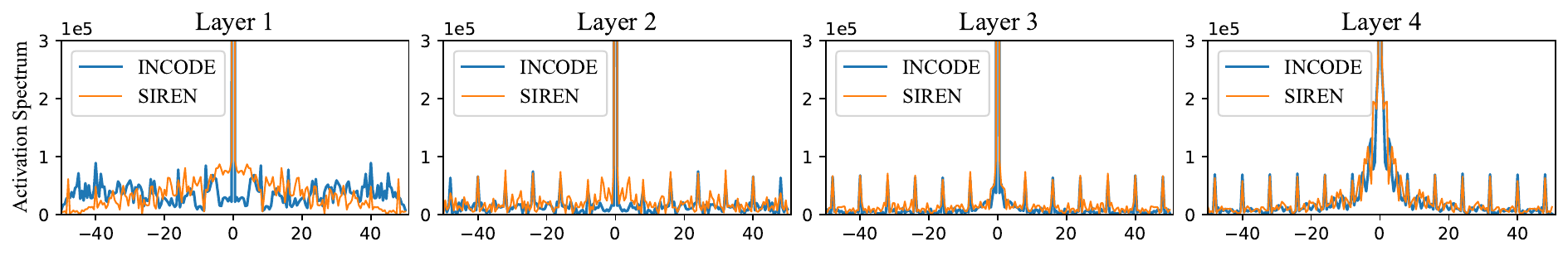}
    \vspace{-1em}
    \caption{Comparison of frequency response representations of the proposed method vs. SIREN across network layers.}
    \label{fig:activation}
    \vspace{-1.25em}
\end{figure*}
\vspace{-1em}
\subsubsection{Composer Network}
\vspace{-0.5em}
We define the composer network as an MLP with $L$ hidden layers, each containing $P$ hidden features that map the input coordinates to its output domain, e.g., RGB values for an image. Each layer within the network utilizes a periodic-nonlinear activation function, which post-activation layer can be defined as follows:
\begin{equation}
    \text{y}_l = \underline{\textbf{a}} \, \sin(\underline{\textbf{b}} \, w_o\left(\text{W}_l \text{y}_{l-1} + \text{b}_{l} \right) + \underline{\textbf{c}}) + \underline{\textbf{d}},
\end{equation}
where values of \underline{\textbf{a}}, \underline{\textbf{b}}, \underline{\textbf{c}}, and \underline{\textbf{d}} are adaptively determined at each iteration throughout the learning process, facilitated by the harmonizer network. In this function, each variable plays a distinct role in shaping the behavior of the activation function. We now proceed to individually analyze the effects of each variable and see how they increase the representation accuracy and robustness:

\noindent\textbf{a (Amplitude)}: The amplitude \underline{\textbf{a}} plays a pivotal role in vertically scaling or stretching the sinusoidal wave. Specifically, our approach involves adjusting \underline{\textbf{a}} during the learning process to influence the strength of the activation function's response. In denoising, a higher \underline{\textbf{a}} could potentially amplify noise, whereas, in representation, it could enhance the emphasis on certain features. As a result, guiding the model to achieve an optimal balance for \underline{\textbf{a}} leads to a feature enhancement in the representation tasks and noise suppression in the denoising-related tasks.
Hence, the first objective of the harmonizer module is to optimally set the \underline{\textbf{a}} value based on the given task. This adaptive learning is a departure from fixed activation functions and allows the model to self-regulate its response based on the specific characteristics of the data.

\noindent\textbf{b (Frequency Scaling)}: The variable \underline{\textbf{b}} governs the frequency scaling of the sinusoidal wave. The adjustment of \underline{\textbf{b}} plays a significant role in accentuating either finer or coarser details within the representation. Thus, careful selection of \underline{\textbf{b}} enables attenuating high-frequency noise in denoising tasks and regulates the granularity of captured features in representation tasks. Hence, adaptive calibration of \underline{\textbf{b}} during the learning process effectively enhances the representation capacity of representation models while reducing the high-frequency noise in the denoising tasks.

\noindent\textbf{c (Phase Shift)}: The phase shift parameter \underline{\textbf{c}} horizontally displaces the sinusoidal wave along the x-axis. This adjustment impacts the alignment of features represented by the activation function, influencing their spatial arrangement within the model's generated representation. Consequently, modifying \underline{\textbf{c}} holds the potential to affect the quality and fidelity of the resulting representation. In denoising, altering \underline{\textbf{c}} can shift noise patterns, altering their perceptibility in the output; therefore, the model can learn to balance the effect of noise by shifting the sinusoidal wave.

\noindent\textbf{d (Vertical Shift)}: The variable \underline{\textbf{d}} in the activation function acts as a vertical shift. Increasing \underline{\textbf{d}} adds a constant positive offset to the entire function, resulting in a raised baseline. This adjustment effectively enhances the overall brightness of the generated image, akin to intensifying light or color. By elevating \underline{\textbf{d}}, the output values of the activation function shift upwards, creating a visually brighter appearance in the representation. Thus, manipulating \underline{\textbf{d}} provides a mechanism for controlling baseline brightness within the INR framework.
Therefore, devising a mechanism that dynamically adjusts these variables during the learning process at each iteration can guide us towards achieving our objectives of constructing a resilient model with substantial representation capacity.
\vspace{-1em}
\subsubsection{Harmonizer Network}
\vspace{-0.75em}
The harmonizer network employs an MLP architecture consisting of $K$ hidden layers and $p_1, p_2, ..., p_K$ hidden features. Its primary function revolves around the direct regulation of the amplitude, frequency, and displacement of the sinusoidal activation such that it is defined as $g(\text{z}; \theta): \mathbb{R}^r\mathbb \rightarrow \mathbb{R}^4$. This network is structured to predict the \textbf{a}, \textbf{b}, \textbf{c}, and \textbf{d} values dynamically. It initiates its function by receiving a latent code $z$ from a task-specific pre-trained model. Subsequently, it endeavors to predict these variable values in an adaptive manner. In our architectural framework, we strategically integrate a task-specific pretrained model, harnessing the invaluable insights gained from its extensive training on a large-scale dataset. This integration is pivotal in dynamically transforming the data into a meaningful latent space, subsequently facilitating its utilization by the harmonizer network.
\vspace{-0.5em}
\subsection{Loss function}
\vspace{-0.75em}
In our approach, we employ the mean squared error (MSE) as a metric to minimize the differences between the predicted signal values and their corresponding true values. This optimization objective aims to make the predicted values closely align with the actual data. Additionally, we introduce a regularization term to the loss function. This term is designed to enforce positive values for the parameters \textbf{a}, \textbf{b}, \textbf{c}, and \textbf{d}. We enforce the variables to be positive in order to guide the model towards more relevant solutions, encourage the model to converge more rapidly, and reduce the likelihood of becoming trapped in a local optimum during the training process. This regularization mechanism contributes to a more efficient and effective optimization process, enhancing the overall performance of the model. Our loss function is defined as follows:
\begin{equation}
\begin{aligned}
    \underset{\theta, \mathbf{a}, \mathbf{b}, \mathbf{c}, \mathbf{d}}{\text{arg min}} \quad & \mathbb{E} \left[ \| f(\mathbf{x}; \theta) - S(\mathbf{x}) \|_{2}^{2} \right] \\
    \textrm{s.t.} \quad & \mathbf{a} \geqslant  1, \quad \mathbf{b} \geqslant  1, \quad \mathbf{c} \geqslant  0, \quad \mathbf{d} \geqslant  0.
\end{aligned}
\end{equation}

We control the strength of the regularization applied to the parameters \textbf{a}, \textbf{b}, \textbf{c}, and \textbf{d} through corresponding coefficients $\lambda_1$, $\lambda_2$, $\lambda_3$, and $\lambda_4$ in the optimization process, enabling us to manage the trade-off between fitting the data and imposing constraints on the parameter values.
\vspace{-0.25em}
\subsection{Expressiveness of INCODE}
\vspace{-0.25em}
This section explores INCODE's expressive capabilities and compares them with the SIREN architecture. Yüce et al. \cite{yuce2022structured} analyze the two-layer SIREN. In a SIREN with two layers and input $x$, the first layer produces $Z^{(0)} = \sin(\Omega x)$, and the second layer yields $Z^{(1)} = \sin(\omega^{(1)} \sin(\Omega x))$. The second-layer output in SIREN can be expressed as:
\begin{equation}
    \sum_{m=0}^{P-1} \sum_{s_{1}, \ldots, s_{N} = -\infty}^{+\infty} \left( \prod_{t=0}^{N-1} J_{s_t} W^{(1)}_{m,t} \right) \sin \left( \sum_{t=0}^{N-1} s_{t}w_{t}x \right),
\end{equation}
where $J_{s}$ defines the Bessel function of the first kind of order $s$. The decreasing nature of $J_{s_t} W^{(1)}_{m,t}$ results in higher-order harmonics carrying smaller weights, concentrating energy around a narrow band centered at input frequencies $\Omega$. Scaling coefficients like $\omega^{(1)}$ amplify higher-order harmonics, enabling a broader range of learnable frequencies.

INCODE introduces a harmonizer network learning activation function parameters $a$, $b$, $c$, and $d$, leading to $a \sin(b \Omega + c) + d$. The simplified second-layer output in INCODE, with only $a$ and $b$, becomes:
\begin{equation}
    a \sum_{m=0}^{P-1} \sum_{s_{1}, \ldots, s_{N} = -\infty}^{+\infty} \left( \prod_{t=0}^{N-1} J_{s_t} W^{(1)}_{m,t} a b \right) \sin \left( \sum_{t=0}^{N-1} s_{t}w_{t} b x \right).
\end{equation}

The term $a$ enhances noise robustness and emphasizes signal details, while $ab$ amplifies coefficients for higher-order terms, broadening the frequency spectrum beyond that of SIREN, given that $ab \geqslant 1$. To ensure this condition, we consider \underline{\textbf{a}} and \underline{\textbf{b}} as $e^{a}$ and $e^{b}$, respectively, in our proposed activation function to fulfill this condition. Parameter $c$ crucially produces $e^{jc}$ terms, effectively controlling $b$ to prevent unbounded growth. This control enhances network stability and maintains meaningful frequency components.

For experimental purposes, an image is generated within specific frequency ranges, transmitting information from low to high frequency. Empirical evidence in \autoref{fig:activation} demonstrates higher amplitudes at higher frequencies in INCODE's first layer, confirming enhanced mapping capabilities and expanded frequency bandwidth compared to SIREN. The parameterization of the harmonizer network achieves broader frequency coverage while retaining sensitivity to essential signal details.
\vspace{-0.75em}
\section{Experiments}
\vspace{-0.6em}
\begin{figure*}[t]
    \centering
    \includegraphics[width=\textwidth]{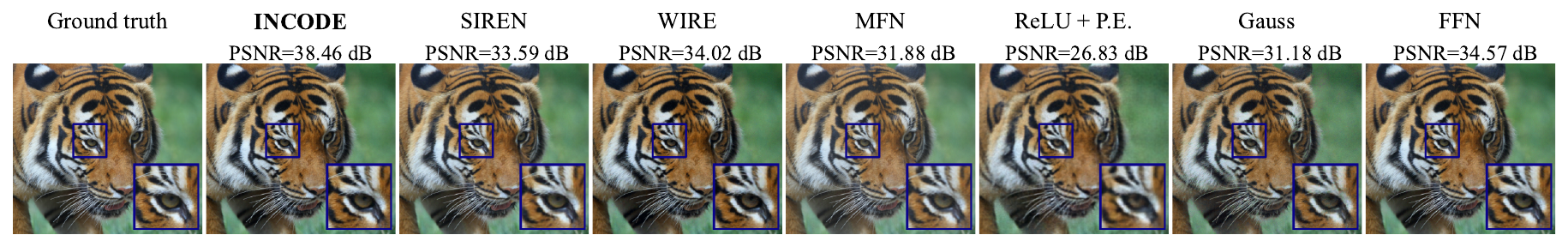}
    \vspace{-1.75em}
     \caption{\textbf{Image representation:} Comparison of INCODE with SOTA methods.}
     \vspace{-0.75em}
    \label{fig:img_exp}
\end{figure*}
\begin{figure*}[!tbh]
    \centering
    \includegraphics[width=\textwidth]{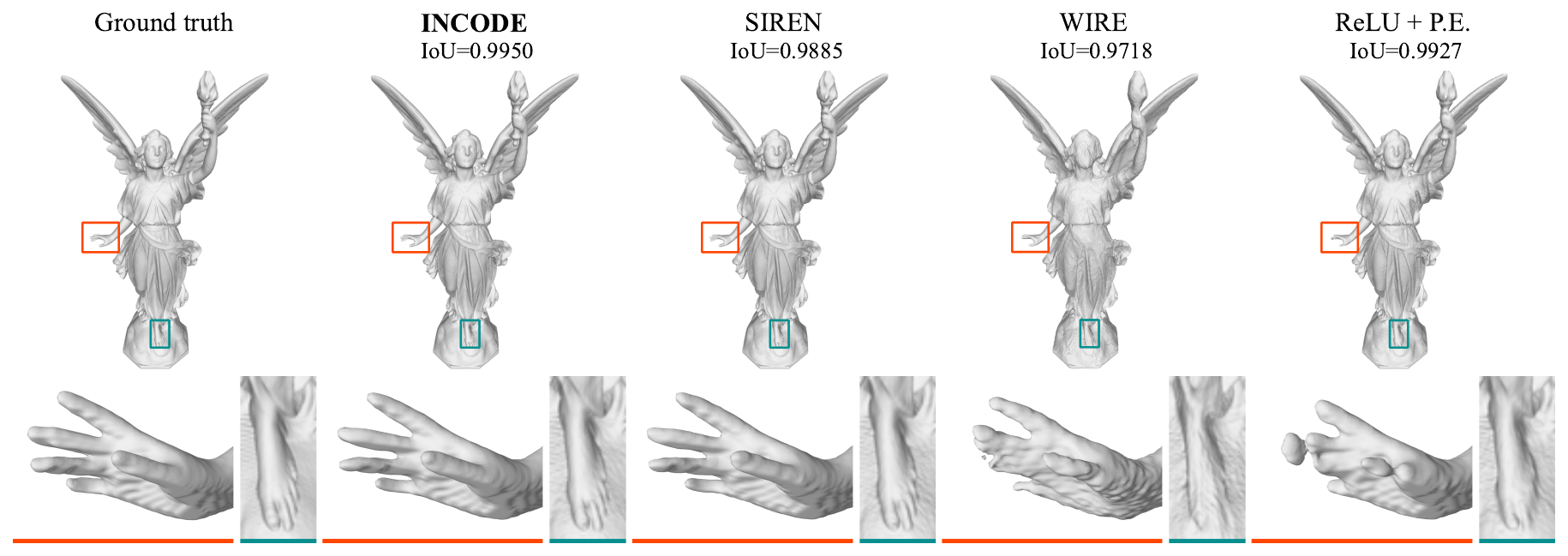}
    \vspace{-1.75em}
    \caption{\textbf{Occupancy volume representation:} Comparison of INCODE with SOTA methods.}
    \vspace{-0.75em}
    \label{fig:sdf_exp}
\end{figure*}
\begin{figure*}[!tbh]
    \centering
    \begin{subfigure}{0.59\textwidth}
        \includegraphics[width=\textwidth]{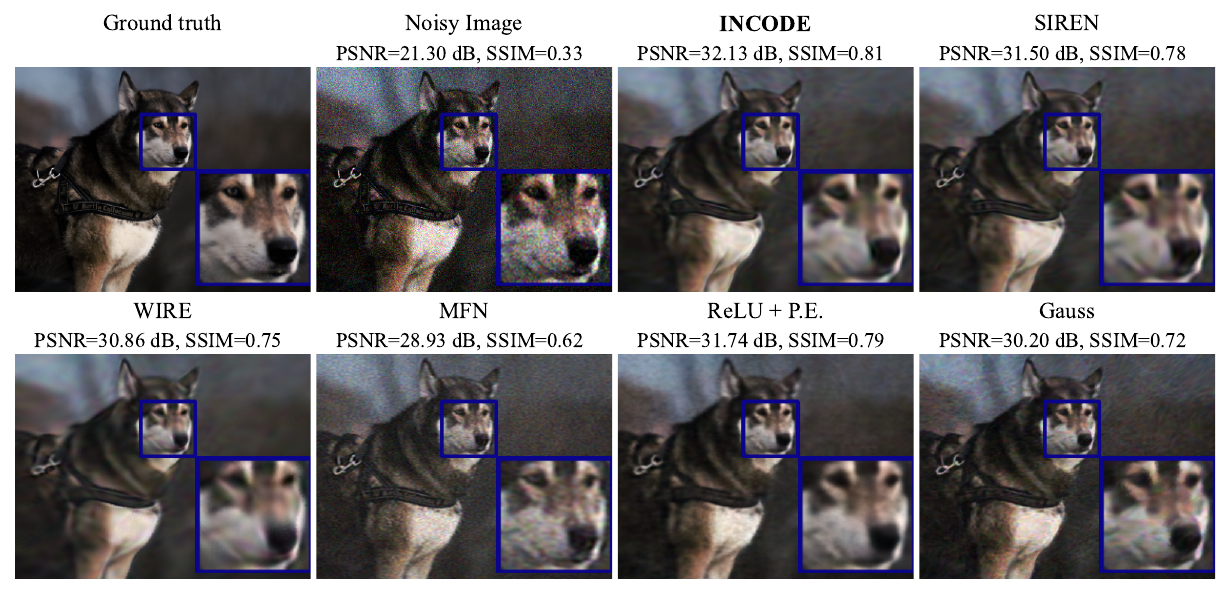}
    \end{subfigure}
    \begin{subfigure}{0.4\textwidth}
        \includegraphics[width=\linewidth]{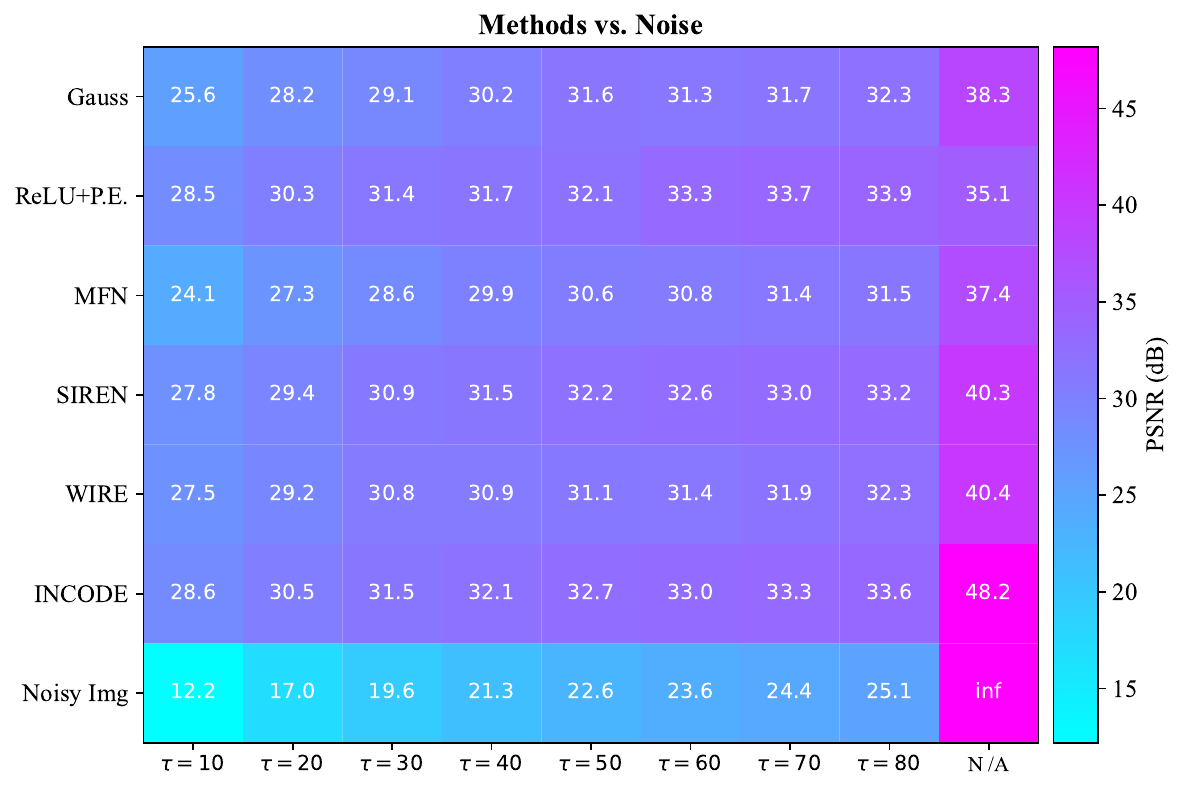}
    \end{subfigure}
    \vspace{-0.5em}
    \caption{\textbf{Image denoising:} Qualitative and quantitative comparison of INCODE with SOTA methods.}
    \vspace{-0.75em}
    \label{fig:exp_denoising_sample}
\end{figure*}
\begin{figure*}[!tbh]
    \includegraphics[width=\textwidth]{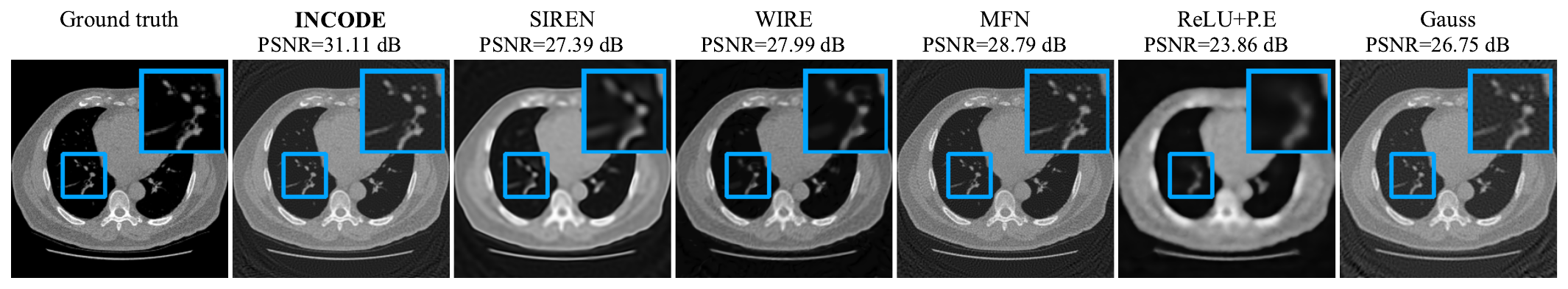}
    \vspace{-1.5em}
    \caption{\textbf{CT Reconstruction:} Comparison of  CT-based reconstruction with 150 angles with SOTA methods.}
    \label{fig:exp_CT}
    \vspace{-1.25em}
\end{figure*}
\noindent\textbf{Implementation Details.} We utilize a 5-layer composer network with 256 units for all experiments. Specifics regarding the harmonizer network can be found in the relevant task description. Our experiments are performed using PyTorch on an Nvidia RTX 3070 Ti GPU with 8GB memory. We use the Adam optimizer~\cite{adam} with a learning rate scheduler, aiding convergence by decreasing the learning rate by $\alpha$ at each epoch's completion. Experiments are conducted for 500 epochs, except audio (1000 epochs), occupancy (200 epochs), and CT reconstruction (2000 epochs). $\lambda_1$, $\lambda_2$, $\lambda_3$, and $\lambda_4$ are set to 0.1993, 0.0196, 0.0588, and 0.0269, respectively, obtained by training the model for the image representation task on 10 samples and using Optuna \cite{optuna_2019} for hyperparameter optimization. We extensively compare our methods with WIRE \cite{saragadam2023wire}, SIREN \cite{saragadam2023wire}, MFN \cite{fathony2021multiplicative}, Gaussian \cite{ramasinghe2022beyond}, ReLU with Positional-Encoding (ReLU+P.E.) \cite{tancik2020fourier}, and FFN \cite{tancik2020fourier}. Further architectural details of these methods are available in the supplementary materials. 
\vspace{-0.25em}
\subsection{Signal Representations}
\vspace{-0.5em}
\subsubsection{Image}
\vspace{-0.5em}
\noindent\textbf{Data.} We conducted our image representation experiments using the DIV2K dataset \cite{Timofte_2018_CVPR_Workshops}, which was downsampled by a factor of 1/4. For example, \autoref{fig:img_exp} is downsampled from $1644 \times2040 \times3$ to $411 \times 510 \times 3$.

\noindent\textbf{Architecture.} The Composer network maps 2D coordinates to RGB values, and $w_o$ is set to 30 for this network. The Harmonizer network is a 3-layer MLP with 64, 32, and 4 features, equipped with the SiLU \cite{elfwing2018sigmoid} activation function. It maps the generated latent code with $r=64$ to the four parameters of activation. Weights of this network are normally initialized $\mathcal{N}(0, 0.001)$ with constant biases of 0.31. We use ResNet34 \cite{he2016deep} truncated to its fifth layer, followed by an adaptive average pooling, to generate the latent code. The learning rate is set to $9\times 10^{-4}$ and $\alpha$ to 0.1. 

\noindent\textbf{Analysis.} The experimental results of image representation are presented in \autoref{fig:img_exp}. The results clearly indicate that INCODE outperforms its counterparts in terms of representation quality. Notably, it achieves a substantial enhancement of +3.89 dB in PSNR values compared to the nearest counterpart, FFN, and +4.44 dB and +4.48 dB improvements compared to WIRE and SIREN, respectively. Additionally, INCODE has shown a sharper reconstruction of the tiger eyebrow than the other methods, particularly ReLU+P.E. and MFN. This observation underscores the promising potential of INCODE for image representation, capable of producing sharper images with finer details. More results are in the supplementary file.
\vspace{-1em}
\subsubsection{Occupancy Volume}
\vspace{-0.5em}
\noindent\textbf{Data.} We use the Lucy dataset from the Stanford 3D Scanning Repository and follow the WIRE strategy \cite{saragadam2023wire}. We create an occupancy volume through point sampling on a $512 \times 512 \times 512$ grid, assigning values of 1 to voxels within the object and 0 to voxels outside.

\noindent\textbf{Architecture.} Our network and training configurations resemble the image representation task, with the distinction that the composer network now maps 3D ($M=3$) coordinates to signed distance function (SDF) values ($N=1$). Utilizing ResNet3D-18 \cite{tran2018closer} truncated to the third layer for feature extraction to generate a latent code of size 128, our approach effectively incorporates volumetric data into the composer network. 

\noindent\textbf{Analysis.} The results showcased in \autoref{fig:sdf_exp} underscore INCODE's effectiveness as a robust replacement for its counterparts in occupancy representation tasks. Remarkably, INCODE adeptly harnesses the informative latent code to condition the composer network, yielding an amplified representation capacity. This augmentation is particularly evident in the intensification of high-frequency information while also adeptly capturing low-frequency details. Our method yields higher Intersection over Union (IOU) values, particularly excelling in replicating intricate details such as Lucy's hand and foot. INCODE remarkably enhances object details and scene complexity, enabling more accurate representation compared to existing methods.
\vspace{-1em}
\subsubsection{Audio Representations}
\vspace{-0.5em}
\noindent\textbf{Data.} We use the first 7 seconds of Bach's Cello Suite No. 1: Prelude \cite{sitzmann2020implicit}, with a sampling rate of 44100 Hz as our example for the audio representation task.

\noindent\textbf{Architecture.} The composer network transforms 1D $(M=1)$ input to its corresponding 1D output $(N=1)$. It employs strategic frequency initialization for effective learning due to the nature of audio: $w_0$ is set to 3000 for the first layer to capture high spatial frequency information, and hidden $w_0$ is set to 30 for subsequent layers. To capture audio features, Mel Frequency Cepstral Coefficients (MFCCs) \cite{logan2000mel} serve as the feature extractor. MFCCs encode both frequency and temporal information, suited for audio representation. The harmonizer network utilizes extracted features and generates the activation parameters. Also, the learning rate is $9\times 10^{-5}$, and $\alpha$ is 0.2.

\noindent\textbf{Analysis.}
We evaluate INCODE's performance against established methods to gauge its effectiveness in audio signal representation. Results highlight INCODE's substantial reduction in error rates and increase of +10.60 dB PSNR value compared to the second best, Guass (See Supplementary, \autoref{fig:exp_audio}). The periodicity of audio signals at various time scales leads to an accurate and efficient representation in INCODE, akin to SIREN. INCODE converges swiftly to a distortion-minimized representation, while Gauss and ReLU+P.E. methods manifest distortion during playback. Although SIREN strives to mitigate this, some dominant noise is witnessed in the background. INCODE notably excels in this aspect, as evidenced in its error figure.

\subsection{Inverse Problems}
\subsubsection{Image denoising}
\vspace{-0.5em}
\noindent\textbf{Data.} We employ an image from DIV2K dataset \cite{Timofte_2018_CVPR_Workshops},  downsampled by a factor of 1/4 from $1152 \times2040 \times3$ to $288 \times 510 \times 3$. We create the noisy image using realistic sensor measurement with readout and photon noise, where independent Poisson random variables are applied to each pixel. The mean photon count ($\tau$) varied between 10 and 80, while the readout count ($ro$) set fixed at 2.

\noindent\textbf{Architecture.} The composer network is similar to previous tasks, however, we set $w_0$ to 10 for the first layer, while the other layers remain at 30. The choice of $w_0$ in the initial layer plays a crucial role in achieving a denoised image with higher fidelity and fewer artifacts. By setting $w_0$ to a lower value, the network becomes more adept at capturing low-frequency information and smoothing out noise-related variations. The first layer $w_0$ can also be calibrated in alignment with the noise characteristics to attain optimal signal quality. The harmonizer network is a 4-layer MLP, containing 32, 16, 8, and 4 nodes. Each layer is followed by a LayerNorm and SiLU activation function. This network is responsible for mapping the latent code ($r=64$) generated by ResNet34 to the activation parameters. Weights are initialized using the normal distribution of $\mathcal{N}(0, 0.001)$ and constant biases of 0.0005. This initialization emphasizes the relevant signal components and suppresses noise-related artifacts. We train the model with a learning rate of $1.5\times 10^{-4}$ and $\alpha=0.1$.

\noindent\textbf{Analysis.} We demonstrate the effectiveness of INCODE in solving inverse problems using the example of image denoising, capitalizing on its inductive bias and robustness. The visual comparison of our approach is presented in \autoref{fig:exp_denoising_sample} for $\tau=40$ and $ro=2$, where INCODE significantly enhances the fidelity of the noisy image with a +10.83 dB PSNR improvement and a 0.48 increment in the Structural Similarity Index (SSIM) metric. INCODE adeptly preserves image details while mitigating noise artifacts, particularly when compared to the MFN and Gauss methods, where noise effects still persist in the output. Furthermore, our approach outperforms the ReLU+P.E. method by 0.39 dB and 0.02 in terms of SSIM. Furthermore, we present a histogram visualization in \autoref{fig:exp_denoising_sample}, wherein the image is subjected to varying degrees of noise to illustrate the comparative performance of each approach and the incremental trend as noise influence diminishes. Additionally, we conduct the methods without noise to exhibit the capacity of each approach in both denoising and representation tasks. Evidently, INCODE has shown a comparable performance with the ReLU+P.E. method, while the ReLU-based networks are particularly good for prioritizing learning low-frequency information, demonstrating the robustness and power of the INCODE in denoising tasks. 
\vspace{-1em}
\subsubsection{Image super resolution}
\vspace{-0.5em}
\noindent\textbf{Data.} We adopt an image from the DIV2K dataset \cite{Timofte_2018_CVPR_Workshops} and downsampled the image with the size of $1356 \times 2040 \times 3$ by factors of 1/2, 1/4, and 1/6.

\noindent\textbf{Architecture.} We maintain the same architectural and training settings as the image representation task. By employing a downsampled image during training, we exploit the interpolation capabilities of INRs to reconstruct an image of its original size in the test.

\noindent\textbf{Analysis.} In super-resolution, the application of INRs as interpolants presents a promising avenue. This notion indicates that INRs possess inherent advantageous biases that can be harnessed to enhance super-resolution tasks. To validate this proposition, we conducted \(1\times\), \(2\times\), \(4\times\), and \(6\times\) super-resolution experiments on an image. As presented in \autoref{tab:exp_sr_results}, the results demonstrate that INCODE consistently achieves superior PSNR and SSIM values across all super-resolution levels, outperforming alternative methods. Furthermore, the visual demonstration of INCODE's superiority is presented in (Supplementary, \autoref{fig:exp_sr_sample}), revealing its ability to retain sharper details compared to others that often result in blurrier outcomes.

\begin{table}[tbh]
    \centering
    \vspace{-0.5em}
    \caption{INCODE vs. SOTAs in super-resolution.}
    \vspace{-0.5em}
    \resizebox{1\linewidth}{!}{
    \begin{tabular}{lcclcclcclcc} 
    \toprule
    \multirow{2}{*}{Methods} & \multicolumn{2}{c}{1$\times$} &  & \multicolumn{2}{c}{2$\times$} &  & \multicolumn{2}{c}{4$\times$} &  & \multicolumn{2}{c}{6$\times$} \\
     & PSNR & SSIM &  & PSNR~ ~ & SSIM &  & PSNR & SSIM &  & PSNR & SSIM \\ 
    \midrule
    Gauss & 30.25 & 0.79 &  & 29.05 & 0.86 &  & 27.07 & 0.83 &  & 25.04 & 0.79 \\
    FFN & 32.96 & 0.90 &  & 29.33 & 0.85 &  & 29.41 & 0.86 &  & 27.01 & 0.84 \\
    MFN & 31.20 & 0.84 &  & 30.78 & 0.87 &  & 29.35 & 0.86 &  & 27.23 & 0.83 \\
    ReLU P.E. & 32.41 & 0.87 &  & 30.29 & 0.88 &  & 25.52 & 0.82 &  & 24.26 & 0.81 \\
    WIRE & 31.57 & 0.84 &  & 31.37 & 0.86 &  & 28.55 & 0.83 &  & 24.77 & 0.72 \\
    SIREN & 32.10 & 0.87 &  & 31.51 & 0.89 &  & 28.81 & 0.85 &  & 26.46 & 0.84 \\
    \midrule
    \textbf{INCODE}{\cellcolor[rgb]{0.686,1,1}} & \textbf{33.44}{\cellcolor[rgb]{0.686,1,1}} & \textbf{0.91}{\cellcolor[rgb]{0.686,1,1}} & {\cellcolor[rgb]{0.686,1,1}} & \textbf{33.02}{\cellcolor[rgb]{0.686,1,1}} & \textbf{0.92}{\cellcolor[rgb]{0.686,1,1}} & {\cellcolor[rgb]{0.686,1,1}} & {\cellcolor[rgb]{0.686,1,1}}\textbf{29.88} & {\cellcolor[rgb]{0.686,1,1}} \textbf{0.87} & {\cellcolor[rgb]{0.686,1,1}} & \textbf{\textbf{27.57}} {\cellcolor[rgb]{0.686,1,1}} & \textbf{0.85} {\cellcolor[rgb]{0.686,1,1}} \\
    \bottomrule
    \end{tabular}
    }
    \vspace{-1.5em}
    \label{tab:exp_sr_results}
\end{table}
\vspace{-0.75em}
\subsubsection{CT reconstruction}
\vspace{-0.5em}
\noindent\textbf{Data.} We use a publicly available CT lung image ($256 \times 256$) from the Kaggle Lung Nodule Analysis dataset \cite{azad2019bi} to assess our model's performance in CT reconstruction.

\noindent\textbf{Architecture.} The architecture remains consistent with that of the image representation task. INCODE involves using the ResNet34 architecture to process the undersampled sinogram and generate a latent code. We conduct training over 2000 epochs, using the learning rate of $2\times 10^{-4}$, coupled with $\alpha=0.4$. We generate a sinogram according to the projection level using the radon transform. The model predicts a reconstructed CT image. Subsequently, we calculate the radon transform for the generated output and compute the loss function between these sinograms, to guide the model toward generating CT images with reduced artifacts.

\noindent\textbf{Analysis.} CT reconstruction is the process of generating a computed image from sensor measurements. Sparse CT reconstruction deals with the added complexity of generating accurate images when only a subset of measurements is available, posing challenges due to limited data constraints. INCODE addresses this challenge by employing a conditional harmonizer network to seamlessly integrate deep prior information into the model. As shown in \autoref{fig:exp_CT}, INCODE stands out by producing sharp reconstructions with clear details using 150 measurements (+2.32 dB improvement compared to the second best, MFN). Conversely, MFN shows artifacts similar to WIRE, and Gauss, yet achieving higher PSNR values. On the other hand, SIREN and ReLU+P.E. yield overly blurred results with reduced details. This underscores INCODE's robustness in addressing challenges posed by noisy and undersampled inverse problems. Its ability to balance image fidelity and noise reduction establishes INCODE as a promising solution in the underconstrained image reconstruction landscape. We also explore the relationship between the number of projections and the reconstructed CT quality (see Supplementary, \autoref{fig:exp_CT_all_projection}). Our method maintained its superiority compared to SOTA methods, underscoring its resilience in the face of measurement noise. 
\vspace{-1em}
\subsubsection{Inpainting}
\vspace{-0.5em}
\noindent\textbf{Data.} W utilize Celtic spiral knots image with a resolution of $572\times 582\times 3$. The sampling mask is generated randomly, with an average of 20\% of pixels being sampled.

\noindent\textbf{Architecture.} We adopt a configuration similar to that of image representation architecture, albeit with adjustments tailored to the task-specific model. Due to the random pixel sampling, a pre-trained model like ResNet cannot be employed. Hence, a custom model is crafted for latent code generation, consisting of two layers [Conv1D, ReLU, MaxPooling], followed by another Conv1D layer. The resulting latent code is of size 64. For training, the model employs a learning rate of $1.5 \times 10^{-4}$ and $\alpha=0.25$.

\noindent\textbf{Analysis.}
Despite only sampling 20\% of pixels, INCODE effectively addresses inverse problems as demonstrated by single-image inpainting. The output in (Supplementary, \autoref{fig:exp_inpainting}) illustrates that INCODE can achieve performance on par but better with other baseline methods. Specifically, INCODE exhibits the ability to generate sharper results with more details.

\subsection{Neural radiance fields}

Neural Radiance Fields (NeRFs) \cite{mildenhall2020nerf} combine INRs and volume rendering by using MLPs equipped with ReLU+P.E, aiming to implicitly represent scenes for synthesizing novel views. By training a 3D implicit function using spatial coordinates $(x, y, z)$ and viewing directions $(\theta, \phi)$, NeRFs can predict the color and density of that specific location. This allows for generating new views of objects from different angles by tracing camera rays through pixels using neural rendering. We, therefore, investigate the effectiveness of using INCODE without positional encoding in the NeRF. We found that our approach yields superior results in fewer epochs. We substantiate the excellence of our approach through comparative analyses and results showcased in the supplementary material.

\section{Conclusion}

In this paper, we have presented INCODE, a transformative approach to Implicit Neural Representations (INRs) that significantly enhances their representation capacity. By introducing a dynamic sinusoidal-based activation function with adaptive control, INCODE overcomes the limitations of existing INRs. The harmonizer network, guided by deep prior knowledge, dynamically adjusts activation function parameters, enabling the model to adapt to specific data characteristics. Our experiments demonstrate the superior performance of INCODE across a wide range of tasks. 

\section{Supplementary Material}
\appendix

\section{Experimental Results}
In this section, we broaden our experimental scope to encompass a more comprehensive comparison between our approach and state-of-the-art (SOTA) methods. We have demonstrated that the inherent simplicity of INCODE contributes to enhanced performance compared to its counterpart SOTA methods, specifically in terms of expressiveness and representation capacity. These findings underscore the efficacy of our approach in pushing the boundaries of INR networks and facilitating their applicability across diverse domains. We now present additional visualizations that distinctly show the advantage of our approach.

\subsection{Image representation}
\begin{figure*}[!t]
    \centering
    \includegraphics[width=\textwidth]{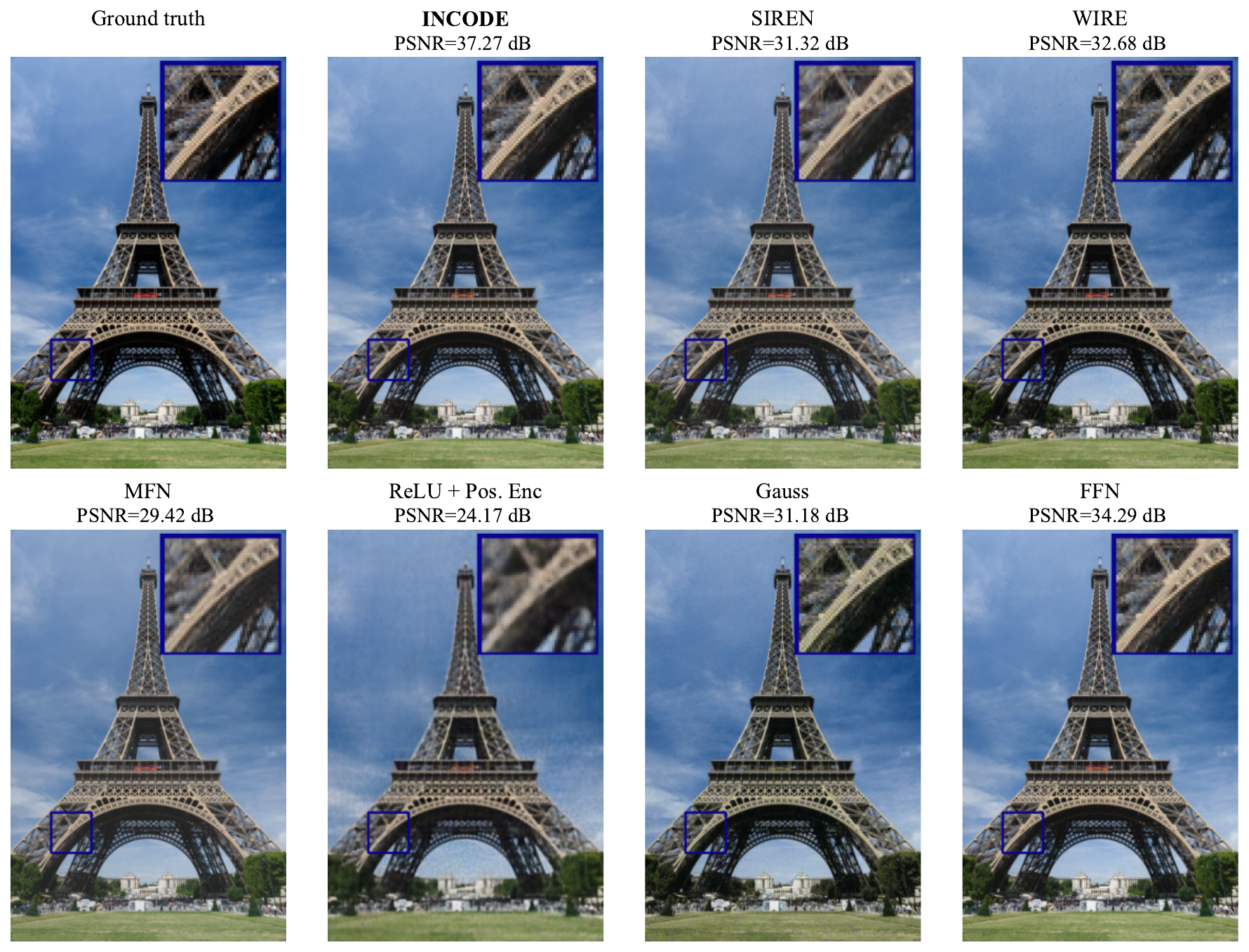}
    \vspace{-0.5em}
    \caption{\textbf{Image representation:} Comparison of INCODE with SOTA methods.}
    \label{fig:img_rep_sample_10}
    \vspace{-0.5em}
\end{figure*}

\begin{figure*}[!thb]
    \centering
    \includegraphics[width=\textwidth]{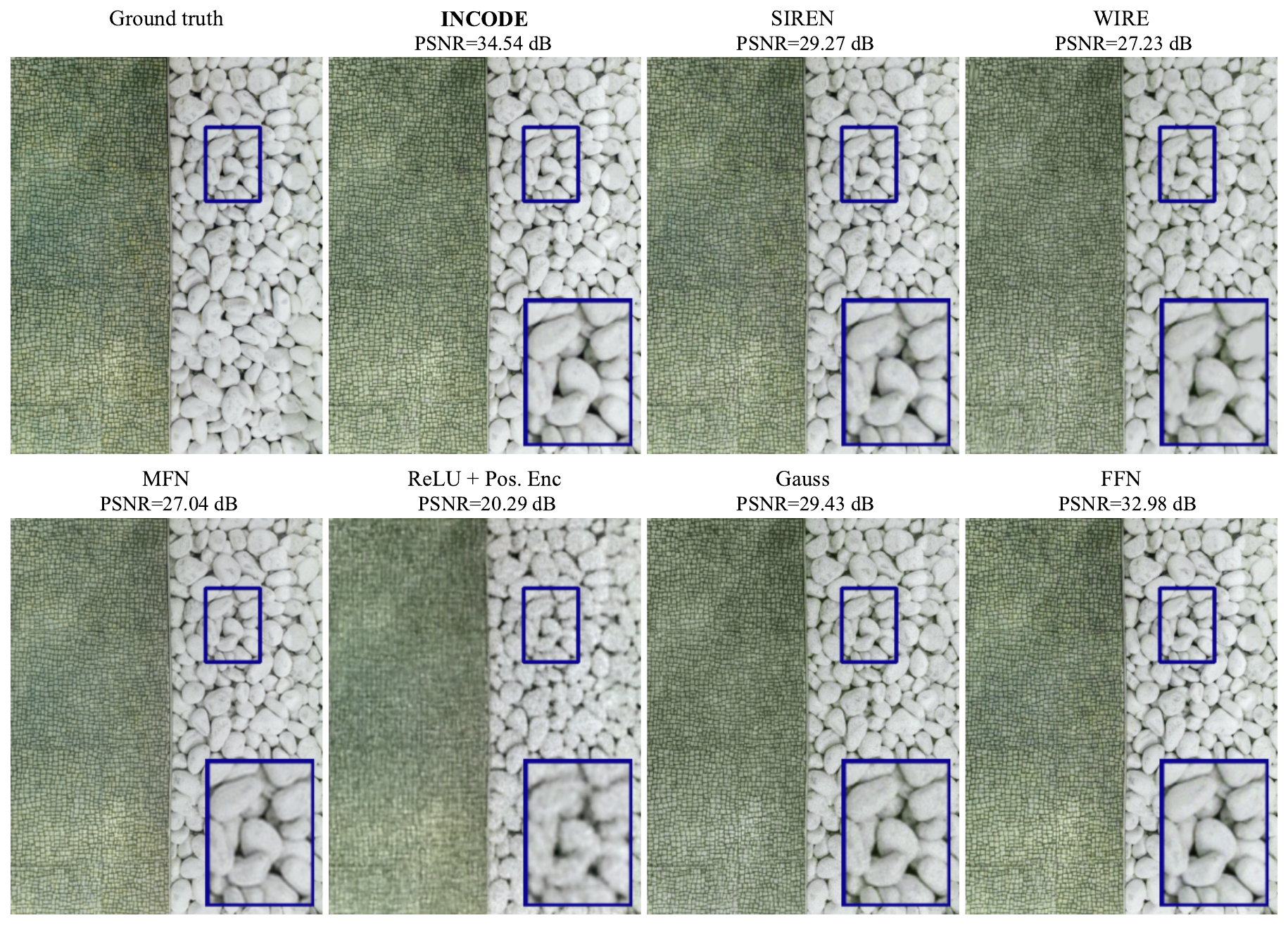}
    \caption{\textbf{Image representation:} Comparison of INCODE with SOTA methods.}
    \label{fig:img_rep_sample_1}
\end{figure*}

As depicted in \autoref{fig:img_rep_sample_10} and \autoref{fig:img_rep_sample_1}, it is evident that INCODE achieves superior qualitative and quantitative performance. Particularly in \autoref{fig:img_rep_sample_10}, INCODE exhibits an approximate accuracy improvement of +2.98 dB and +4.59 dB compared to FFN \cite{tancik2020fourier} and WIRE \cite{saragadam2023wire}, respectively. The zoomed-in image distinctly illustrates INCODE's ability to grasp intricate details of the Eiffel Tower. In contrast, ReLU+P.E. and MFN \cite{fathony2021multiplicative} yield blurry outcomes, while Gauss \cite{ramasinghe2022beyond} displays slight color alteration, although it captures certain intricate features. Gauss also struggles to recognize the orange object positioned at the tower's center. Likewise, SIREN \cite{sitzmann2020implicit} fails to capture the full complexity of the tower's structure, leading to a smoothed and blurred representation.

Additionally, \autoref{fig:img_rep_sample_1} presents a challenging image with intricate patterns, posing a challenge for representation. Notably, INCODE and FFN emerge as the sole methods achieving a PSNR value over 30 dB, with INCODE exhibiting a +1.56 improvement over the second-ranking FFN. As evidenced in the zoom-in image, ReLU+P.E. expectedly yields a blurred output, given the inherent properties of its ReLU activation function. Interestingly, WIRE and Gauss encounter difficulty in precisely grasping the image's color characteristics, leading to slight color differences. While MFN effectively addresses this color challenge, it falls short in capturing the image's intricate details, particularly its edges.

Overall, our study shows that INCODE excels in image representations. It consistently outperforms other methods across various images, even with intricate patterns. This success is due to INCODE's ability to capture intricate details. While alternative methods faced challenges in representing complex patterns, colors, or high-frequency information, INCODE exhibited competence in addressing these challenges. Thus, our findings highlight INCODE as one of the optimal choices for robust and superior image representation.
\begin{figure*}[h]
    \centering
    \includegraphics[width=0.9\textwidth]{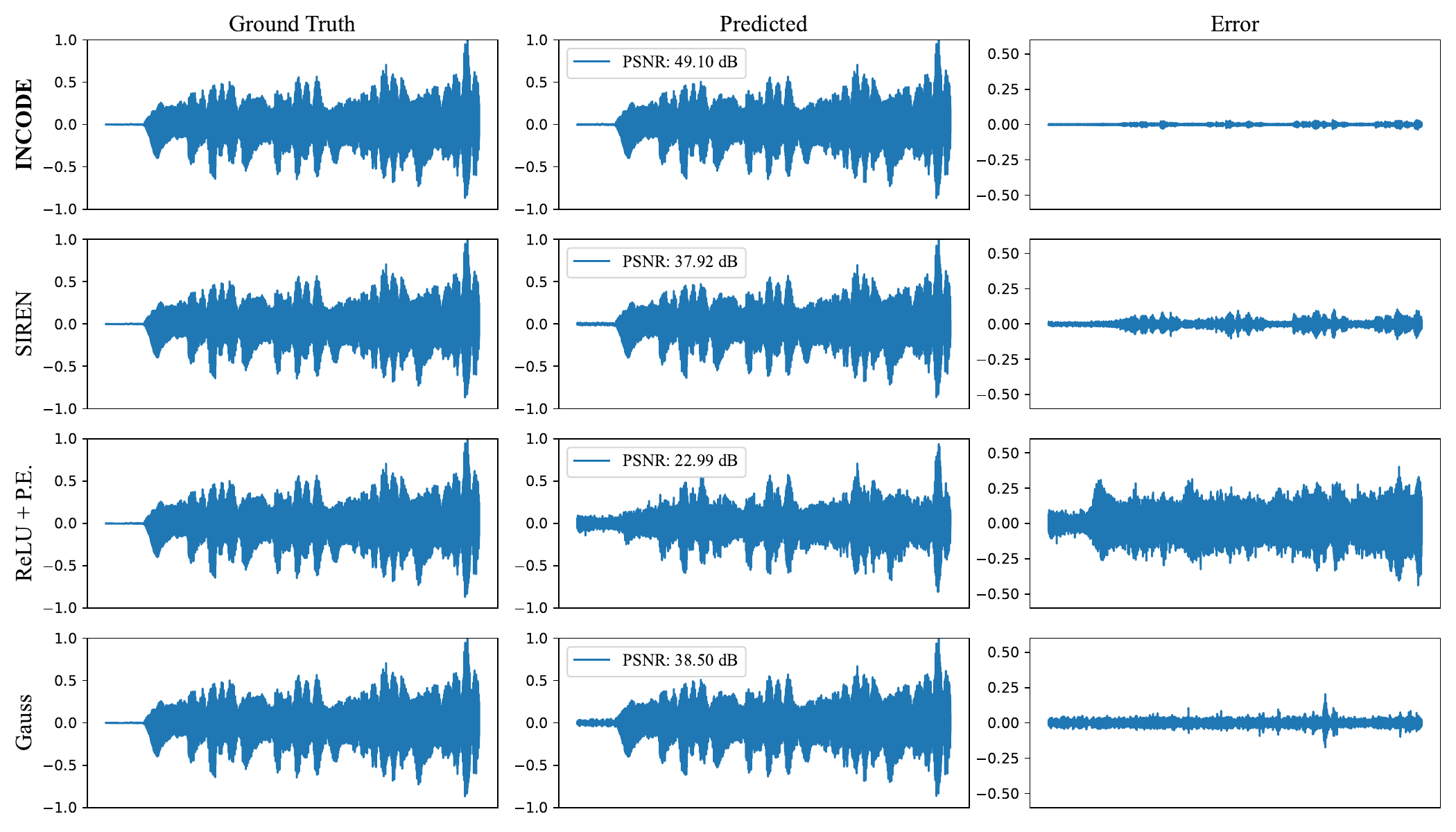}
    \vspace{-0.75em}
    \caption{\textbf{Audio representation:} We compare INCODE with SOTA methods for audio representation. In the third column, we display the reconstruction error.}
    \vspace{-1em}
    \label{fig:exp_audio}
\end{figure*}
\subsection{Audio representation}
We present audio representation visualization results along with its error maps in \autoref{fig:exp_audio}. These visualizations help to understand the strength of our approach. We have provided a detailed analysis of these results in the main section of the paper to ensure a comprehensive understanding of our findings. In terms of sound playback quality, Gauss introduces a noticeable squeak-like sound that accompanies the main audio. With ReLU+P.E., noise dominance becomes more pronounced, making it difficult to discern the original sound. While employing SIREN, some moments are marred by bothersome noise, as indicated by the error map. However, INCODE significantly outperforms these methods by having notably less noise interference. This aspect positions INCODE as a favorable choice for encoding audio data with improved quality.

\subsection{Super resolution}
\begin{figure*}[!t]
    \centering
    \includegraphics[width=\textwidth]{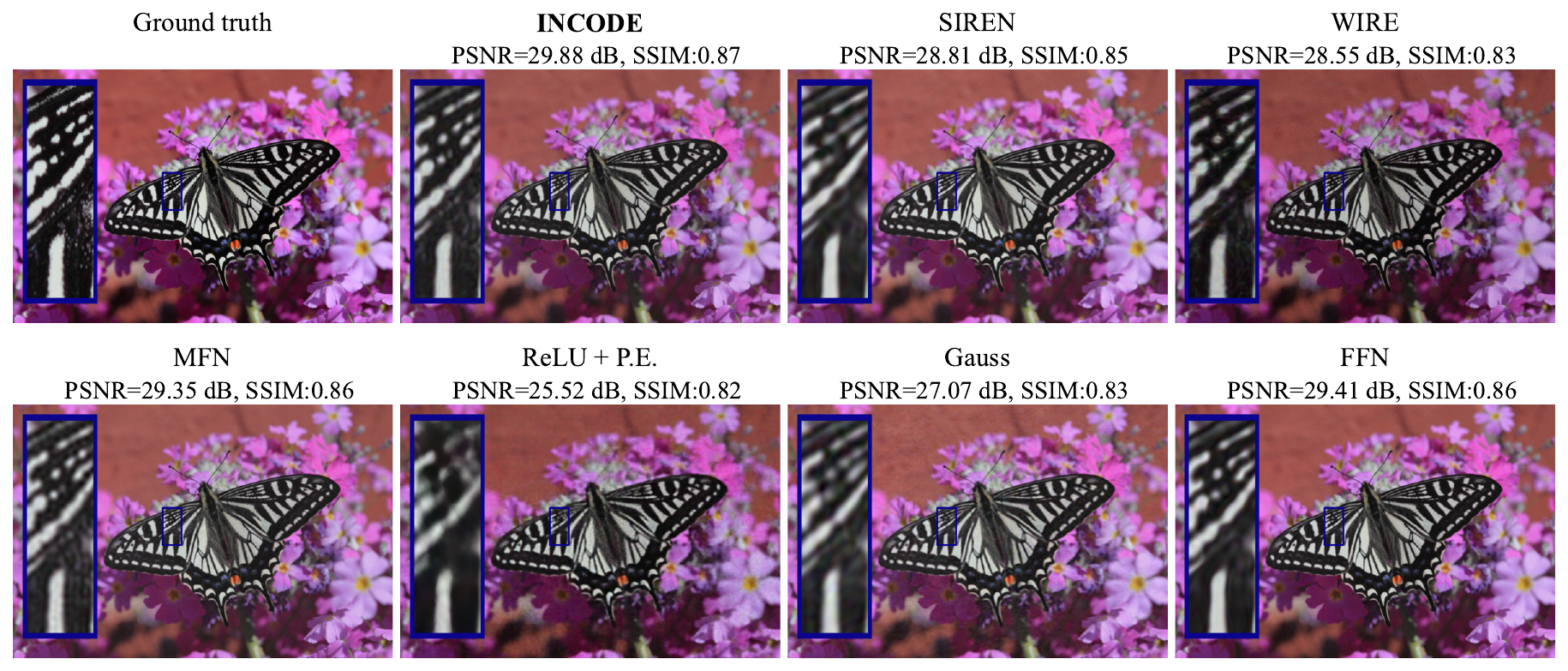}
    \caption{\textbf{Super Resolution. } Results of a 4× single image super-resolution using various approaches }
    \label{fig:exp_sr_sample}
\end{figure*}

To illustrate the efficacy of our approach in the super-resolution task, we have included a visual comparison of 4$\times$ super-resolution in \autoref{fig:exp_sr_sample}. From a quality perspective, INCODE produces sharper results with finer details in the butterfly's wing, while the blurred outcomes of SIREN, FFN, Gauss, and ReLU+P.E. are evident,  even though the quantitative values are relatively close. This visual comparison supports our quantitative findings in \autoref{tab:exp_sr_results} (see the main paper) and affirms INCODE's proficiency in super-resolution tasks, where it offers better quality when performing upsampling.

\begin{figure}[t]
\centering
    \includegraphics[width=\linewidth]{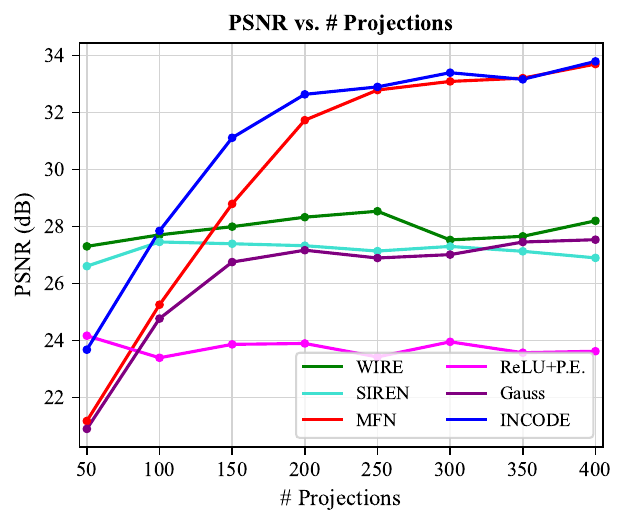}
    \caption{INCODE vs. SOTAs in CT reconstruction across different numbers of projections.}
    \vspace{-1em}
    \label{fig:exp_CT_all_projection}
\end{figure}
\subsection{Computed Tomography (CT) reconstruction}
Under-measurement in CT samples results from a range of factors that reduce the accuracy of the imaging process. Artifacts, stemming from issues like patient movement during scanning, metallic objects causing beam distortion, and equipment calibration problems, contribute to discrepancies. INRs address these concerns and solve this inverse problem by leveraging their inductive bias. We investigate the impact of varying the number of measurements (ranging from 50 to 400, with increments of 50) as shown in \autoref{fig:exp_CT_all_projection}. Notably, SIREN, WIRE, and ReLU+P.E. yield consistent results across all measurements. Particularly, WIRE excels in CT reconstruction with 50 measurements; however, increasing the data information in such models doesn't enhance their performance, indicating saturation. In contrast, INCODE exhibits considerable improvement as measurements increase from 100 to 400, showcasing the effectiveness of incorporating deep prior information. Notably, INCODE with 150 measurements outperforms all nonlinearities in the full range of projection numbers, except for MFN, which closely competes after reaching 200 projections and performs the second best. These findings acknowledge the robustness and power of INCODE in addressing under-measurement challenges within CT reconstruction.

\subsection{Inpainting}
\begin{figure}[!tbh]
\centering
    \includegraphics[width=0.95\linewidth]{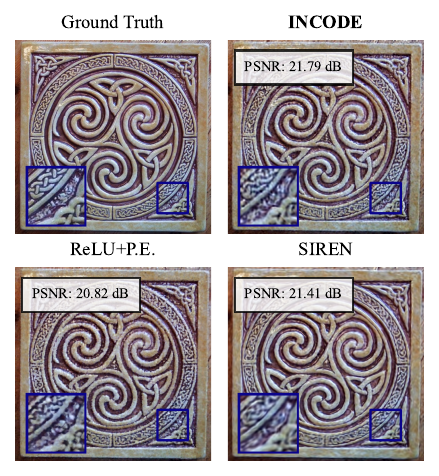}
    \caption{\textbf{Image inpainting:} Comparison of INCODE with SOTA methods.}
    \label{fig:exp_inpainting}
\end{figure}

Image inpainting poses a formidable challenge as models are tasked with predicting entire pixel values based on only a fraction of trained pixel data. The high capacity of INR provides the opportunity to accomplish this inverse problem challenge. The strong prior ingrained within the space of INR functions paves the way for applications like inpainting from limited observations, where it uses the learned representation of the trained model to predict inpainting missing values. Our approach involves randomly sampling 20\% of the pixels and then employing the model's learned representation to predict the missing pixels. The comparison result is shown in \autoref{fig:exp_inpainting}. As observed in other tasks, INCODE's power in capturing intricate features, particularly edges, stands out compared to other methods that tend to yield blurred outcomes. While a modest +0.38 dB improvement is noted compared to SIREN, the visual presentation demonstrates that SIREN, much like ReLU+P.E., struggles to comprehensively capture high-frequency details.


\begin{figure*}[!tbh]
\centering
    \includegraphics[width=\textwidth]{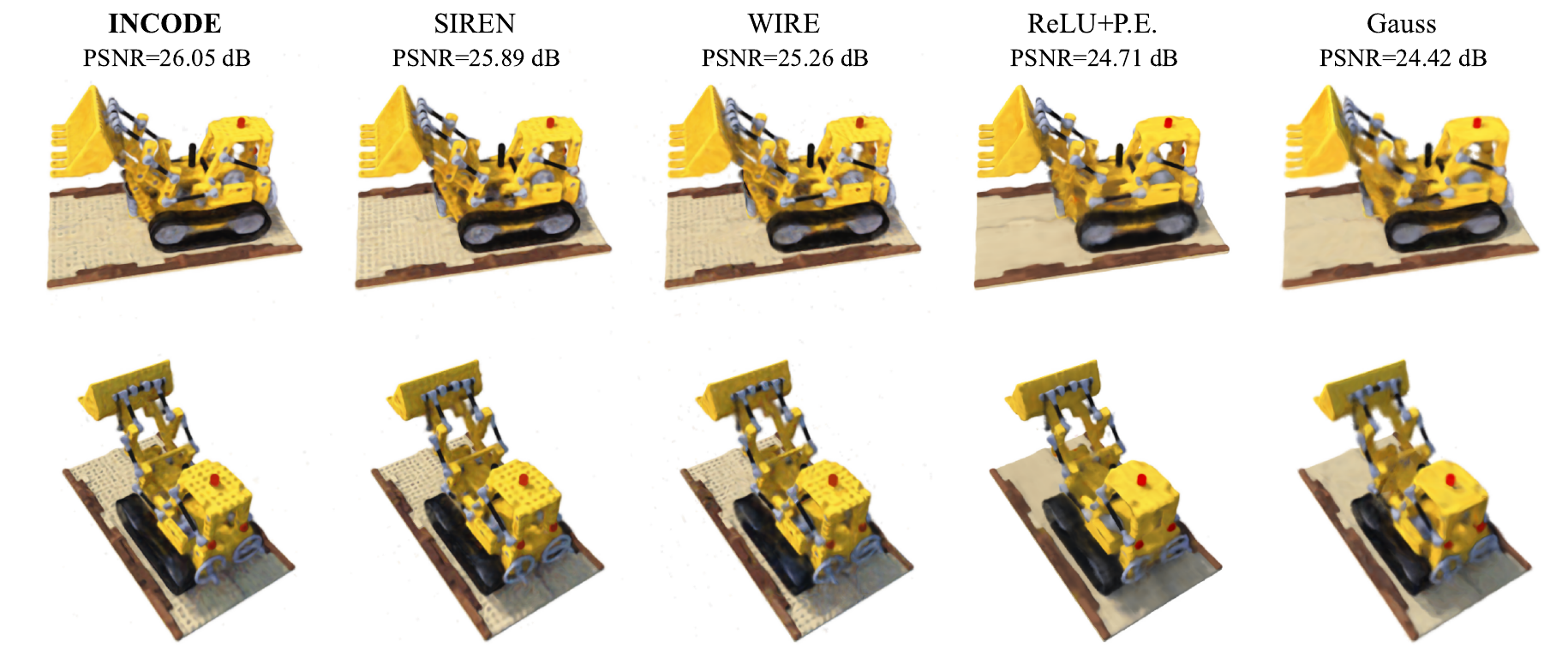}
    \caption{\textbf{Neural Radiance Fields:} The figure presented above illustrates rendered images generated by a neural radiance field using different methods. Notably, INCODE consistently outperforms all other methods in terms of visual reconstruction quality, highlighting its robust feature representation.}
    \label{fig:exp_nerf}
\end{figure*}
\subsection{Neural radiance fields}
In our approach, we followed a strategy akin to \cite{saragadam2023wire}, making use of the publicly available torch-ngp package \cite{tang2022torch,tang2022compressible} to train the NeRF model. Our NeRF architecture encompasses two main networks: one for predicting sigma ($\sigma$) and the other for determining color ($RGB$). These networks are constructed as 4-layer MLPs, each with 182 hidden features.

Additionally, we introduced two harmonizer networks, one for the sigma network and another for the color network. These harmonizers employ 4-layer MLPs, featuring 32, 16, 8, and 4 nodes, with each layer followed by LayerNorm and the SiLU activation function. They receive a latent code and condition their corresponding composer networks, which are initialized similarly to the denoising task.

To generate the latent code, we utilized a truncated ResNet34 model at its fifth layer, followed by adaptive average pooling. During training, a single random image from the training dataset was used, and for testing and validation, we again employed one random training image. The color MLP took positional coordinates $(x, y, z)$ and direction parameters $(\theta, \phi)$ as inputs, while the sigma MLP solely required positional information.

For our experimental results, depicted in \autoref{fig:exp_nerf}, we utilized a Lego dataset comprising 100 training images, each downsampled by 1/2 to $400\times400$ dimensions, for training the NeRF. Subsequently, we evaluated the model's performance on an additional 200 images. Training of the NeRF models was conducted on an A-100 GPU with 20 GB of memory. Throughout training, we used learning rates of $3 \times 10^{-4}$ for INCODE, $3 \times 10^{-4}$ for SIREN, $6 \times 10^{-4}$ for WIRE, $3 \times 10^{-3}$ for Gauss, and $1 \times 10^{-2}$ for ReLU+P.E. The learning rate is decreased to $0.1\times$ initial value over a total of 3000 training epochs to achieve their optimal outputs. Additionally, we set omega  ($\omega_0$) to 40 for INCODE, SIREN, and WIRE, and sigma ($s_0$) to 40 for WIRE and Gauss. Apart from ReLU, we did not use positional encoding for other nonlinearities to highlight their individual capabilities.

As shown in \autoref{fig:exp_nerf}, our approach achieves a +0.16 dB improvement over SIREN and a +0.79 improvement compared to WIRE. Qualitative results also demonstrate a superior performance of INCODE compared to SOTA models. Notably, INCODE excels in capturing fine-grained details and information. For instance, it effectively captures intricate features such as the middle black connector in the loader, while SIREN failed to learn. Also, INCODE outperforms other methods like WIRE, ReLU+P.E., and Gauss, which exhibit blurred and smooth results in comparison.

\begin{figure*}[!tbh]
\centering
    \includegraphics[width=\textwidth]{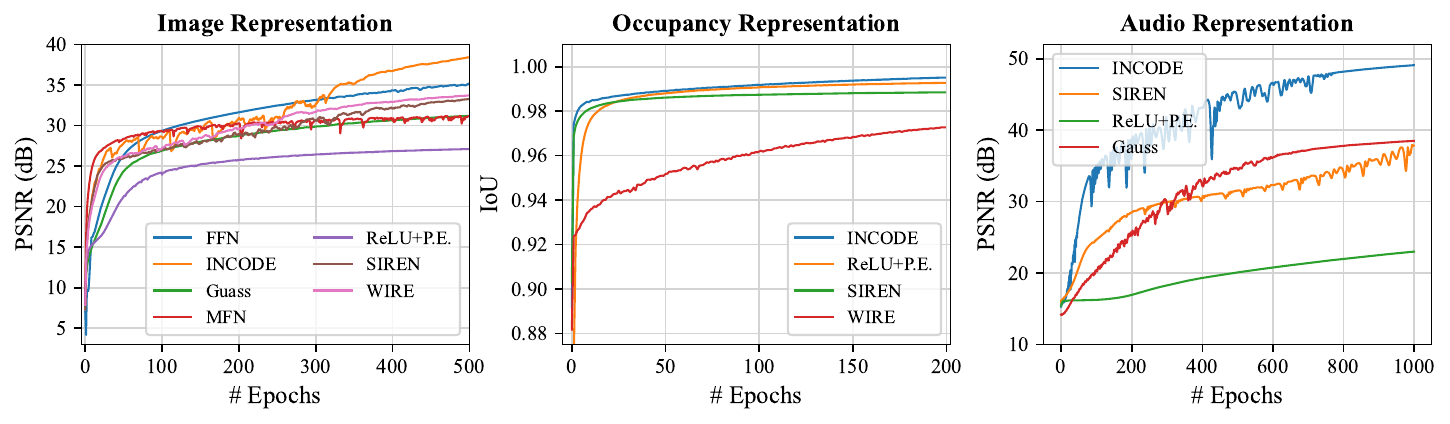}
    \caption{\textbf{Convergence rates in different representations:} Explore the convergence rates of Image, Occupancy volume, and Audio representations. }
    \label{fig:convergance_rate}
\end{figure*}
\section{Experimental Analysis}
\subsection{Convergence rate comparison}
We analyze the convergence rate of INCODE in comparison to other methods across three distinct representation tasks: image, occupancy volume, and audio, as depicted in \autoref{fig:convergance_rate}. The data used for each task corresponds to the respective domain in the main paper.  Remarkably, INCODE consistently showcases accelerated convergence compared to SOTA architectures. This expedited convergence is most pronounced in the audio domain, where a substantial gap between SIREN and INCODE is evident. Leveraging its robust approximation capacity, INCODE achieves fast convergence with high fidelity, rendering it an apt choice for representing different signals.
\begin{figure}[!tbh]
\centering
    \includegraphics[width=\linewidth]{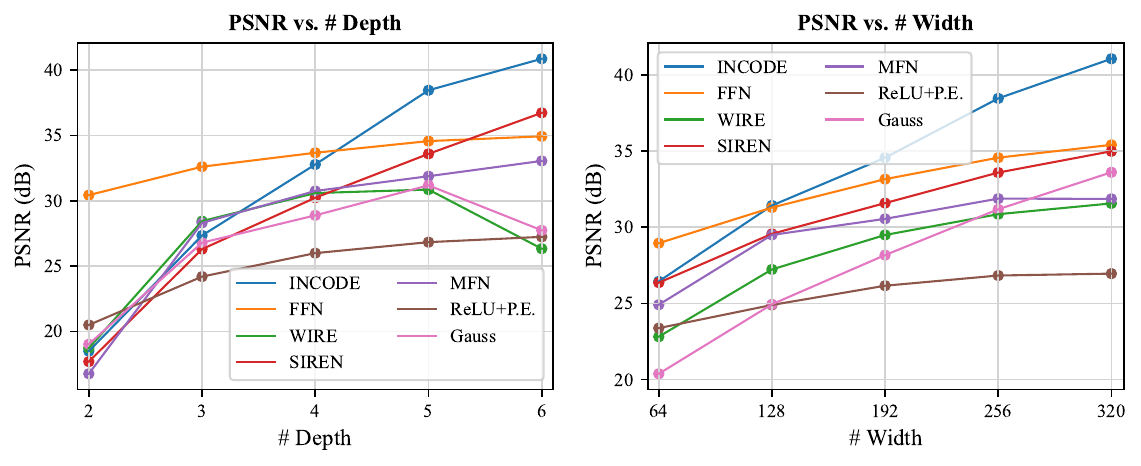}
    \caption{\textbf{Impact of network depth and width:} Explore the influence of network depth and width on performance.}
    \label{fig:depth_width_ablation}
\end{figure}

\subsection{Impact of depth and width of the network}
The analysis of the network's depth and width are presented in \autoref{fig:depth_width_ablation}, which sheds light on the impact of architectural parameters in shaping the performance of INCODE. By systematically varying the number of hidden layers and their width, we gain insights into the trade-off between model complexity and approximation accuracy.

In the left figure, we vary the network's depth from 2-layer MLP to 6-layer, while keeping the width constant at 256. Notably, INCODE exhibits competitive performance compared to other methods in lower layers. However, as the network deepens, INCODE distinctly outperforms FFN, demonstrating its capacity to effectively capture more intricate information with increasing model depth. Shifting to the right figure, we explore the effect of hidden features by adjusting the network's width from 64 to 320, in increments of 64, while maintaining a 5-layer MLP. The trend depicted in the plot accentuates INCODE's remarkable performance, showcasing a steep ascent. Throughout the spectrum of hidden feature counts, INCODE consistently outperforms other SOTA methods. This observation highlights INCODE's proficiency in capturing broader patterns as the width of the network expands, underlining its versatility and ability to adapt to varying levels of complexity.

\section{Experimental details}
In all experiments, we employed a 5-layer MLP with 256 hidden features for all architectures. However, for WIRE, we followed their recommended structures as outlined in their paper to achieve optimal performance. Specifically, for image-based tasks, we used a 4-layer MLP with $s_0=30$ and $\omega_0=20$, featuring 300 hidden features. For the occupancy task, we utilized a 4-layer MLP with 256 hidden features, alongside $s_0=40$ and $\omega_0=10$. In the case of CT reconstruction, we employed a 5-layer MLP with 256 hidden features and set $s_0=10$ and $\omega_0=10$. Lastly, for the denoising task, we opted for the same architecture as the image representation and for $s_0=4$ and $\omega_0=4$. In FFN, a mapping input size of 256 is utilized, for instance, to map image coordinates from 2 to 512, and the parameter $\mathcal{B}$, a random Gaussian matrix, is scaled by a factor of 10. We configured the value of $s_0$ for the Gauss model as follows: $s_0=30$ for image representation, $s_0=100$ for audio representation, and $s_0=10$ for the inverse problem tasks. In addition, we utilized the same initial parameters as described for INCODE in the case of SIREN.

{\small
\bibliographystyle{ieee_fullname}
\bibliography{egbib}
}

\end{document}